\documentclass[11pt,a4paper]{article}
\usepackage[hyperref]{acl2021}
\usepackage{times}
\usepackage{latexsym}
\usepackage{graphicx}
\usepackage{booktabs}

\usepackage{amsmath,bm}
\usepackage{url}
\usepackage{xcolor}
\usepackage{ulem}
\usepackage{multirow}

\usepackage{microtype}

\aclfinalcopy 

\usepackage{booktabs}

\title{Three Sentences Are All You Need: Local Path Enhanced Document Relation Extraction}

\author{
    Quzhe Huang, 
    Shengqi Zhu, 
    Yansong Feng\thanks{\;\;Corresponding author.}~~,  
    Yuan Ye,
    Yuxuan Lai,
    Dongyan Zhao \\
    Wangxuan Institute of Computer Technology, Peking University, China\\
    The MOE Key Laboratory of Computational Linguistics, Peking University, China\\
    {\tt \{huangquzhe, zhusq, fengyansong, pkuyeyuan, erutan, zhaody\}} 
    \\ {\tt @pku.edu.cn} \\
}

\date{}

\begin{document}
\maketitle
\begin{abstract}
Document-level Relation Extraction (RE) is a more challenging task than sentence RE as it often requires reasoning over multiple sentences. Yet, human annotators usually use a small number of sentences to identify the relationship between a given entity pair. In this paper, we present an embarrassingly simple but effective method to heuristically select evidence sentences for document-level RE, which can be easily combined with BiLSTM to achieve 
good performance on benchmark datasets, even better than fancy graph neural network based methods.  We have released our code at https://github.com/AndrewZhe/Three-Sentences-Are-All-You-Need.

\end{abstract}

\section{Introduction}
The task of relation extraction (RE) focuses on extracting relations between entity pairs in texts, and has played an important role in information extraction. While earlier works focus on extracting relations within a sentence \cite{lin2016neural, zhang-etal-2018-graph},
recent studies begin to explore RE at document level \cite{peng2017cross,zeng-etal-2020-double, nan-etal-2020-reasoning},
which is more challenging as it often requires reasoning across multiple sentences.

\begin{figure}
    \centering
    \includegraphics[width=0.96\linewidth,height=0.43\linewidth]{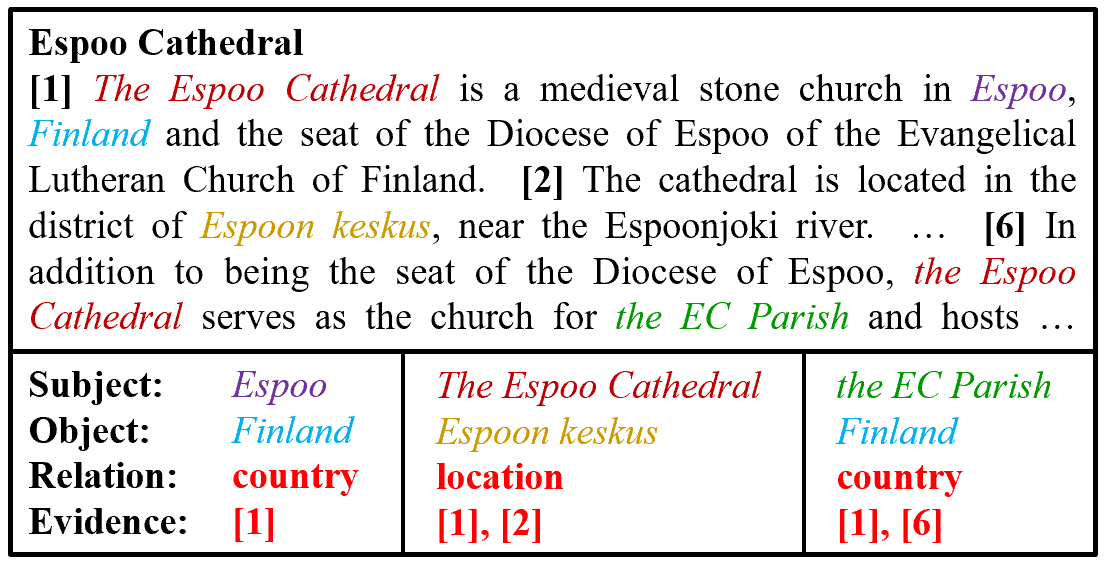}
    \caption{A case extracted from the DocRED dataset. While the document has 6 sentences, only 1 or 2 sentences form the evidence for each relation instance.}
    \label{fig:intro_case}
    \vspace{-0.4cm}
\end{figure}

Compared with sentence level extraction, documents are significantly longer with useful information scattered in a larger scale. However, given a pair of entities, one may only need a few sentences, not the entire document, to infer their relationship; reading the whole document may not be necessary, since it may introduce unrelated information inevitably.
As we can see in Figure~\ref{fig:intro_case}, $S[1]$ is sufficient to recognize \textit{Finland} as the country of \textit{Espoo}, and recognizing the rest two instances requires just 2 sentences as supporting evidence as well. Although the document contains 6 sentences and evidence may span from $S[1] \sim S[6]$, identifying \textit{each} relation instance can be achieved by just reading through 1 or 2 related sentences. This naturally leads us to consider a question: \textit{given an entity pair, how many sentences are required to identify a relationship between them?} We perform a pilot study across 
3 widely-used document RE datasets, DocRED \cite{yao_docred_2019}, CDR \cite{li2016biocreative} and GDA \cite{wu2019renet}. As shown in Table~\ref{tab:size_of_supporting_evidence}, we find that more than $95\%$ instances require no more than 3 sentences as supporting evidence, and $87\%$ even requires only 2 or less.

Our preliminary finding suggests that, instead of taking the entire document as context, a case-specific selection may be more useful to help a model  focus on the most relevant and informative evidence. Previous studies apply graph neural networks (GNNs) for this filtering process~\cite{christopoulou2019connecting,zeng2020double}. Here, GNNs are used to collect relevant information from the entire context through an aggregation scheme~\cite{nan-etal-2020-reasoning} and achieve great performance, but the selection of crucial evidence from documents is still implicit and lacks interpretability. If, as indicated by our pilot study, most entity relationships can be decided with just 1 $\sim$ 3 evidence sentences, is there a simpler method that can filter the document explicitly while maintaining the crucial information?

We take a closer look at how entity pairs are contextually related in the annotated supporting evidence, and find that annotators tend to select sentences that can connect the two entities.
We therefore design three heuristic rules to extract a small set of \textit{paths} from the document, which can be seen as an approximation of the supporting evidence. Specifically, the \textit{Consecutive Paths} consider the scenario where the head and tail entities are close in the context: if they are within 3 consecutive sentences, we regard these sentences as one path. The \textit{Multi-Hop Paths} correspond to the entity pairs in distant sentences, which can be bridged via other entities that co-occur with the head entity and tail entity in different sentences. 
As the third relation in Figure~\ref{fig:intro_case} shows, \textit{Finland} co-occurs with \textit{The Espoo Cathedral} in S[1] and with \textit{the EC Parish} in S[6], which makes it a bridge to connect \textit{The Espoo Cathedral} and \textit{the EC Parish}. In this case, S[1] and S[6] compose a multi-hop path. When neither of the above rules applies, we collect all the pairs of sentences where one contains head entity and the other contains tail entity as \textit{Default Paths}.

By comparing our path set with human-annotated supporting evidence, we find that up to $87.5\%$ of the supporting evidence can be fully covered by our heuristically selected paths. In other words, our straightforward and interpretable rules serve as an effective proxy to select supporting evidence from documents. We further feed our selected paths to a simple neural network model and obtain surprisingly good performance on DocRED, showing that our selected evidence can retain sufficient information from the entire document to support document-level relation extraction. 

\section{Do we need the entire document?}
\label{sec:do_we}

\begin{table}[]
\centering
\setlength{\tabcolsep}{4pt}
\small
\begin{tabular}{lccccc|c}
\toprule
          & 0     & 1                          & 2      & 3     & \textgreater{}=4 & \# Sent \\ \midrule
DocRED     & 3.7\% & \multicolumn{1}{c}{49.7\%} & 34.3\% & 8.4\% & 3.8\%            & 8.0       \\ \midrule
CDR  & 0.0\% & \multicolumn{1}{c}{68.0\%} & 30.0\% & 0.0\% & 2.0\%            & 9.7       \\
GDA  & 0.0\% & 66.0\%                     & 19.0\% & 3.0\% & 5.0\%            & 10.2      \\ \bottomrule
\end{tabular}

\caption{The proportion of instances with different supporting evidence sizes. \# Sent shows the average number of sentences in a document.}
\label{tab:size_of_supporting_evidence}
\vspace{-0.25cm}
\end{table}
 
For document RE, the major challenge is that the subject and object involved in a relationship may appear in different sentences. Thus, more than one sentence is required to capture the relations. 
Nonetheless, how many sentences from the entire document are required to identify the relationship between an entity pair? To address this question, we analyze the supporting evidence presented in DocRED. 
The supporting evidence for a relation instance refers to all the sentences that can be used to decide whether this relation holds between the entity pair,  
labeled by human annotators \cite{yao_docred_2019}. 
Table~\ref{tab:size_of_supporting_evidence} shows the proportions of entity relation instances with different number of supporting sentences. As can be seen, 

more than 96\% of the DocRED instances are associated with at most 3 supporting evidence. These only take up 37.5\% of a document, since the average document length is 8 sentences.
This means that reading a small part of a document is adequate for one to identify an entity relation instance. 

We further extend our study to two widely used document RE datasets, CDR \cite{li2016biocreative} and GDA \cite{wu2019renet}, 
where CDR is manually constructed and GDA is distantly supervised.
In order to find the minimal number of sentences required, we ask annotators to label a minimal set of sentences that are exactly sufficient to identify an entity relation instance, instead of including all relation-associated sentences as the original DocRED pattern. We randomly select 100 instances respectively from CDR and GDA for this further annotation, and the results are shown at the bottom of Table~\ref{tab:size_of_supporting_evidence}\footnote{As GDA is a distantly supervised dataset, 7 instances that are found wrongly labeled are discarded.}. Although the average length of documents in GDA and CDR are longer than DocRED, it turns out that one can still use no more than 3 supporting sentences to identify over 95\% of the entity relation instances.  
The results on CDR and GDA confirm our previous finding that, a very small number of sentences (or more exactly, no more than 3 sentences) would make it sufficient for human annotators to recognize almost all entity relation instances in a document in widely-used benchmark datasets.

\begin{figure*}
    \centering
    \includegraphics[height=0.36\linewidth]{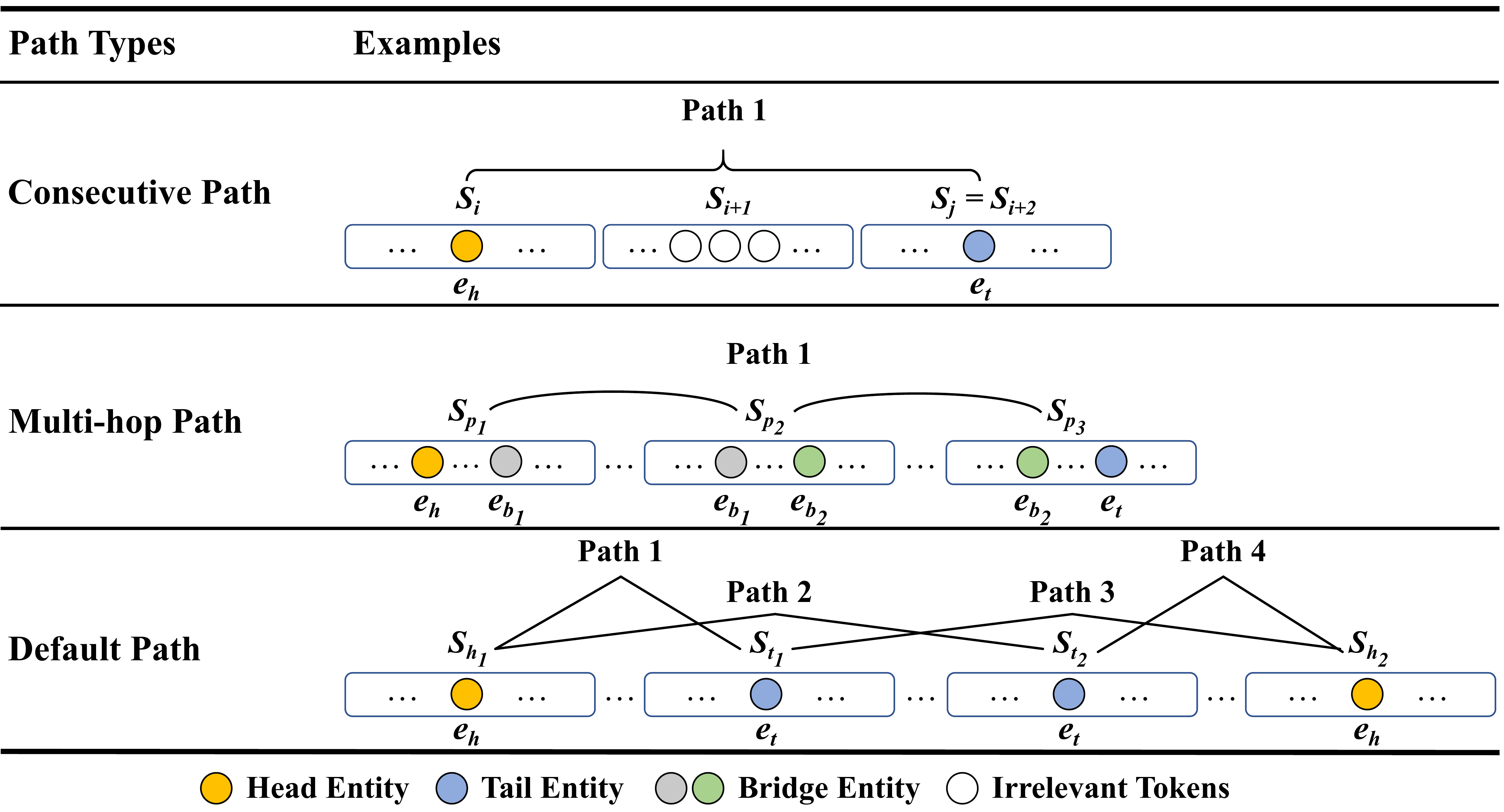}
    \caption{Types of paths connecting head and tail entities. The rounded rectangles represent sentences and the circles are mentions of involved entities or other irrelevant tokens. $e_h$ and $e_t$ stands for a mention of head and tail entities respectively, and $S_*$ represents a sentence.}
    \label{fig:path_example}
\end{figure*}

\section{Which sentences are decisive?}
\label{sec:decisive}

Now our question is how to select the supporting sentences  that are sufficient to identify an entity relation instance.
Intuitively, the supporting evidence should be the sentences that build up the \textit{connection} between a pair of entities. 
Thus, we aim to extract sentence \textit{paths} from the head entity to the tail entity to describe how they are connected. 
As for the simplest case, if there exists one sentence that contains both the head and tail entities, the sentence itself can be seen as a path (the intra-sentence case). For more complex situations where the head and tail entities do not co-occur in one sentence, we define the following 3 types of paths which indicate how the head and tail entities can be possibly related in the context. Figure~\ref{fig:path_example} provides a visualization of the three types of paths.

\paragraph{Consecutive Paths}
Previous studies have shown that the majority of inter-sentence relations are often in nearby text~\cite{swampillai2010inter,quirk-poon-2017-distant}.
We thus select the consecutive sentences to form a path when the head and tail entities are in nearby sentences. Formally, if 
one mention of the head entity appears in sentence $S_i$ and one mention of the tail entity is in sentence $S_j$, these two sentences along with the sentence in between, i.e., sentence $S_{i+1}$, …, $S_{j-1}$ (or $S_{j+1}$, …, $S_{i-1}$ when $i \geq j$) forms a possible path that connects the two entities. 
Given that no more than 3 sentences would suffice for inference, we limit the length of these \textit{Consecutive Paths} to be at most 3, which means $|j-i| \leq 2$.
Note that this definition can be naturally extended to the intra-sentence case where $j=i$. We thus consider the intra-sentence case as a type of the Consecutive Path.
A pair of entities can correspond to multiple consecutive paths since they can be mentioned more than once.

\paragraph{Multi-Hop Paths}
Another typical case for inter-sentence relation instances is the multi-hop relation \cite{yao_docred_2019,zeng-etal-2020-double}. In such cases, the head and tail entities are far from each other in the document but can be connected through \textit{bridge entities}, just like the entity \textit{The Espoo Cathedral} in Figure~\ref{fig:intro_case} bridges \textit{the EC Parish} and \textit{Finland} in sentence $1$ and $6$.

For these cases, we start from the head entity, go through all the bridge entities, arrive at the tail entity, and select all the corresponding sentences in this route as a path. Formally, for the head entity $e_h$ and the tail entity $e_t$, the multi-hop relation indicates that there exist a list of bridge entities $e_{b_1}$, …, $e_{b_k}$ such that $(e_h, e_{b_1})$, $(e_{b_1}, e_{b_2})$, …, $(e_{b_k}, e_t)$ form $k+1$ intra-sentence relations respectively in sentence $S_{p_1}$, …, $S_{p_{k+1}}$. Following this route, we choose these $k+1$ sentences as the \textit{Multi-Hop Path}. Given the discovery in \S\ref{sec:do_we} that most instances only needs 3 sentences, we restrict $k$ to be at most 2, i.e., with only 1 or 2 bridge entities. 
It is possible to have several multi-hop paths for a certain pair with different lists of bridge entities.

\paragraph{Default Paths}~ If neither of the aforementioned rules applies, we consider a rough estimate for the evidence with the most relevant sentences. 
We collect all pairs of sentences where one contains the head entity and the other contains the tail entity as \textit{Default Paths}. Formally, let $\{S_{h_1}, …, S_{h_p}\}$ and $\{S_{t_1}, …, S_{t_q}\}$  denote the sets of sentences that contain the head entity $e_h$ and the tail entity $e_t$, respectively. For this entity pair, we will have $p \times q$ Default Paths $\{S_{h_1}, S_{t_1}\}, …, \{S_{h_p}, S_{t_q}\}$. 
Note that this type of paths is extracted only when no paths are found with the previous two patterns. 

\section{Comparing with Annotated Evidence}

\begin{table}[]
\centering
\small
\setlength{\tabcolsep}{11pt}
\begin{tabular}{lccc}
\toprule 
      & Path Recall & \#Sent & \#Path \\ \midrule
C       & 71.7\%   & 2.31 & 1.71                \\
M       & 31.5\%    & 3.14  & 2.35                      \\
C+M     & 80.5\%    & 2.73  & 2.37                      \\
C+M+D   & 87.5\%    & 2.69 & 2.27                      \\ \midrule
document    & -    & 8.00   & -                  \\ \bottomrule
\end{tabular}
\caption{
C, M and D stand for Consecutive Paths, Multi-hop Paths, and Default Paths, respectively. \#Path and \#Sent are the average path numbers and average sentence numbers in the union of all paths.
} 
\label{tab:overlap}
\vspace{-0.25cm}
\end{table}

To demonstrate the effectiveness of our heuristic rules, we check the size of our path set on DocRED and their consistency with the gold supporting evidence. As mentioned in \S\ref{sec:do_we}, the gold annotation acts as a collection of all related evidence, while each of our extracted paths represents one possible and minimal sentence set. Ideally, if the path set is sufficient, all connecting sentences between the entity pair should be successfully captured. In other words, they would be presented via various paths in our path set. Therefore, the union of paths is expected to be a superset of the supporting evidence. We use the \textbf{Coverage} of the supporting evidence to measure the \textit{sufficiency} of our path set, which stands for the percentage of instances whose supporting evidence is fully covered by the union of our paths. Meanwhile, the total number of paths ($\#Path$) and union size of the paths ($\#Sent$) should also remain at a low standard, so as to avoid \textit{redundancy}.

Table~\ref{tab:overlap} shows the statistics of the path sets extracted via our rules. The Consecutive Paths form a strong baseline that covers $71.7\%$ of instances. Combining the three types, up to $87.5\%$ of instances from the supporting evidence are fully covered by our path sets. 
The main reason that C+M+D can not cover all the instances is that the supporting evidence annotated in DocRED includes all associated sentences, while C+M+D only find a sufficient set to identify the relation.

Meanwhile, notice that the union of the three types contains only $2.69$ different sentences on average, which means that our methods can filter out up to $2/3$ of the original text. Also, our method is computationally efficient since only 2.27 paths need to be modeled on average. This demonstrates that our methods form a sufficient and non-redundant estimate for the gold supporting evidence, drastically alleviating the negative impact of irrelevant information.

\section{Experiments}
\begin{table}[]
\setlength{\tabcolsep}{4pt}
\centering
\small
\begin{tabular}{lccclc}
\toprule
Model       & \multicolumn{3}{c}{Dev}     &  & Test  \\ \cmidrule{2-4} \cmidrule{6-6} 
            & Intra-F1 & Inter-F1 & F1    &  & F1    \\ \midrule
CNN         & 51.87    & 37.58    & 43.45 &  & 42.26 \\
BiLSTM      & 57.05    & 43.49    & 50.94 &  & 51.06 \\
HIN-Glove   & 60.83    & 48.35    & 52.95 &  & 53.30 \\ \midrule
GAT         & 58.14    & 43.94    & 51.44 &  & 49.51 \\ 
GCNN        & 57.78    & 44.11    & 51.52 &  & 51.62 \\
EoG         & 58.90    & 44.60    & 52.15 &  & 51.82 \\
AGGCN       & 58.76    & 45.45    & 52.47 &  & 51.45 \\
LSR-Glove   & 60.83    & 48.35    & 55.17 &  & 54.18 \\
GAIN-Glove  & 61.67    & 48.77    & 55.29 &  & 55.08 \\ \midrule
Paths+BiLSTM & \textbf{62.73}    & \textbf{49.11}    & \textbf{56.54} &  & \textbf{56.23} \\ \bottomrule
\end{tabular}
\caption{Model performance on DocRED.}
\label{tab:nn_res}
\vspace{-0.3cm}
\end{table}

To further validate the sufficiency of our selected paths, we perform evaluation on DocRED  by feeding the paths to an RE model.  While previous works take entire documents as input, we replace the document with our selected paths regarding a given entity pair. Intuitively, if the paths can cover all crucial information in the document, we would expect comparable or better 
performance with identical model architecture, as our paths contain little irrelevant information and may help focus on a few key sentences.

\noindent\textbf{Setup}~
Given a pair of entities, all paths are first extracted as described in \S\ref{sec:decisive}. Since each path corresponds to one possible connection of the head and tail entities, we predict the relations with each path independently and aggregate the results afterwards.

For every single path $c$, we concatenate all sentences in it as one segment $[\mathbf{w_1^c}, ..., \mathbf{w_m^c}]$, where the order of sentences is the same as in the original document. The segment is fed to a BiLSTM to obtain the contextual embeddings $[\mathbf{h_1^c}, ..., \mathbf{h_m^c}]$. The representation of an entity mention, which spans from the $s$-th word to the $t$-th word, is defined as $\mathbf{m_k^c} = \frac{1}{t-s+1} \sum_{j=s}^t \mathbf{h_j^c}$. The representation of an entity $e_i^c$ with $K$ mentions is computed as the average of the representations of its mentions: $\mathbf{e_i^c} = \frac{1}{K} \sum_k \mathbf{m_k^c}$. Then, we use a two-layer perceptron to calculate the probability of each relation $r$ based on the current path $c$: $P^c_{ij}(r) = \sigma(F([e_i^c; e_j^c; |e_i^c - e_j^c|; e_i^c * e_j^c]))$, where $\sigma(\cdot)$ is the Sigmoid function and $F(\cdot)$ stands for the two-layer perceptron.

After obtaining the prediction of every path between a given entity pair, we aggregate the predicted results by selecting the most likely predictions: $P_{ij}(r) = \max_c P^c_{ij}(r)$.

We use the Glove-100 \cite{glove} embedding for the BiLSTM encoder with hidden size 256. Following previous works \cite{nan2020reasoning}, we report the F1 for intra- and inter-sentence entity pairs along with the overall F1 score as evaluation metrics.

\noindent\textbf{Results}~
We compare our methods with previous sequence-based models and graph-based models. All these models take the entire document as input. As shown in Table~\ref{tab:nn_res}, our selected path with BiLSTM achieves $56.23\%$ F1 on the test set, which outperforms the sequence-based models. Compared with the baseline BiLSTM, our model brings $5.68\%$ and $5.62\%$ improvement on intra- and inter-sentence entity pairs on the dev set, respectively. 

Surprisingly, our simple method achieves a higher performance compared with graph-based models, which are more complex and also possess the ability to filter out irrelevant information. Combined with our path-selection scheme, a BiLSTM can perform $1.25\%$ and $1.15\%$ better on the dev and test set, respectively, compared to the SOTA graph-based model in the same situation. This may indicate that, while graph-based models have shown excellent abilities to focus on important information in a self-adaptive manner, it is more helpful to explicitly select from the document than to fully rely on graph-based models. With a simple filtering scheme inspired by human annotations, we can better explore the potentials of existing models and produce better results.

\section{Discussion}
So far we have shown from experiments the limited number of sentences required to deduce a relation instance. While the interesting results seem unconventional for Document RE, which features complex inter-sentence relations, it is worth mentioning that possible explanations exist in current works in related fields. The interdisciplinary outlooks may provide helpful insights for community members to understand the causes of the \textit{three-sentences} phenomenon and revisit the problem of Document-level Relation Extraction.

\paragraph{Linguistic Perspective} 
One likely cause of the discussed phenomenon is that the seemingly distant relations are not so difficult given their linguistic form. \citet{stevenson2006fact} mentions that a majority of inter-sentence relation instances are in fact due to \textit{co-references} (anaphoric expressions or alternative descriptions). In these cases, relations could be considered to be described entirely within one sentence but with head or tail entities being referred to indirectly. Considering anaphoric expressions are likely to appear in surrounding sentences for the candidate mentions~\cite{chowdhury2013controlled}, these findings are directly in line with our observation that consecutive paths could support more than 70\% relation instances, and provide evidence for \textit{three-sentences} phenomenon.

\paragraph{Cognitive Perspective} 
Another possible explanation is that the RE task is naturally defined within a limited amount of entities and context, given the nature of the human brain. It is widely believed that \textit{Working Memory} (WM)~\cite{baddeley1992working} plays a vital role to store and manipulate information in inference tasks~\cite{barreyro2012working}, but the capacity of separate information chunks in WM are often limited to 4~\cite{cowan2001magical}. As we need to memorize all the separate entities in the inference chain along with their relations, 
it is natural that we tend to describe a relation within a limited number of sentences,
since rendering a relationship with more sentences may cause our WM to exceed its capacity. \citet{daneman1980individual} show that the success rate of completing a reading task drastically drops if too much information, exceeding the subject’s WM capacity, is required for the task. Therefore, as the datasets are constructed from natural language, the \textit{three-sentences} phenomenon in the data may be a common pattern that we (unconsciously) follow for mutual understanding.

\section{Conclusion}
In this paper, we perform an analysis over 3 document RE benchmark datasets, and find that human annotators often use a small number of sentences  to extract entity relations in document level. 
This motivates us to think over which sentences are critical for document RE. We carefully design heuristic rules to select informative \textit{path} sets from entire documents, which can be further combined with a simple BiLSTM to achieve competitive performance on a benchmark dataset, even better than complex graph-based methods.

\section*{Acknowledgments}
We thank the anonymous reviewers for the helpful comments and suggestions. This work is supported in part by the National Hi-Tech R\&D Program of China (2018YFC0831900) and the NSFC Grants (No.61672057, 61672058).

\bibliographystyle{acl_natbib}
\bibliography{anthology,acl2021}

\begin{thebibliography}{20}
\expandafter\ifx\csname natexlab\endcsname\relax\def\natexlab#1{#1}\fi

\bibitem[{Baddeley(1992)}]{baddeley1992working}
Alan Baddeley. 1992.
\newblock Working memory.
\newblock \emph{Science}, 255(5044):556--559.

\bibitem[{Barreyro et~al.(2012)Barreyro, Cevasco, Bur{\'\i}n, and
  Marotto}]{barreyro2012working}
Juan~Pablo Barreyro, Jazm{\'\i}n Cevasco, D{\'e}bora Bur{\'\i}n, and
  Carlos~Molinari Marotto. 2012.
\newblock Working memory capacity and individual differences in the making of
  reinstatement and elaborative inferences.
\newblock \emph{The Spanish journal of psychology}, 15(2):471.

\bibitem[{Chowdhury and Zweigenbaum(2013)}]{chowdhury2013controlled}
Md~Faisal~Mahbub Chowdhury and Pierre Zweigenbaum. 2013.
\newblock A controlled greedy supervised approach for co-reference resolution
  on clinical text.
\newblock \emph{Journal of biomedical informatics}, 46(3):506--515.

\bibitem[{Christopoulou et~al.(2019)Christopoulou, Miwa, and
  Ananiadou}]{christopoulou2019connecting}
Fenia Christopoulou, Makoto Miwa, and Sophia Ananiadou. 2019.
\newblock \href {https://doi.org/10.18653/v1/D19-1498} {Connecting the dots:
  Document-level neural relation extraction with edge-oriented graphs}.
\newblock In \emph{Proceedings of the 2019 Conference on Empirical Methods in
  Natural Language Processing and the 9th International Joint Conference on
  Natural Language Processing (EMNLP-IJCNLP)}, pages 4925--4936, Hong Kong,
  China. Association for Computational Linguistics.

\bibitem[{Cowan(2001)}]{cowan2001magical}
Nelson Cowan. 2001.
\newblock The magical number 4 in short-term memory: A reconsideration of
  mental storage capacity.
\newblock \emph{Behavioral and brain sciences}, 24(1):87--114.

\bibitem[{Daneman and Carpenter(1980)}]{daneman1980individual}
Meredyth Daneman and Patricia~A Carpenter. 1980.
\newblock Individual differences in working memory and reading.
\newblock \emph{Journal of verbal learning and verbal behavior},
  19(4):450--466.

\bibitem[{Li et~al.(2016)Li, Sun, Johnson, Sciaky, Wei, Leaman, Davis,
  Mattingly, Wiegers, and Lu}]{li2016biocreative}
Jiao Li, Yueping Sun, Robin~J Johnson, Daniela Sciaky, Chih-Hsuan Wei, Robert
  Leaman, Allan~Peter Davis, Carolyn~J Mattingly, Thomas~C Wiegers, and Zhiyong
  Lu. 2016.
\newblock Biocreative {V CDR} task corpus: A resource for chemical disease
  relation extraction.
\newblock \emph{Database: the journal of biological databases and curation},
  2016:baw068.

\bibitem[{Lin et~al.(2016)Lin, Shen, Liu, Luan, and Sun}]{lin2016neural}
Yankai Lin, Shiqi Shen, Zhiyuan Liu, Huanbo Luan, and Maosong Sun. 2016.
\newblock \href {https://doi.org/10.18653/v1/P16-1200} {Neural relation
  extraction with selective attention over instances}.
\newblock In \emph{Proceedings of the 54th Annual Meeting of the Association
  for Computational Linguistics (Volume 1: Long Papers)}, pages 2124--2133,
  Berlin, Germany. Association for Computational Linguistics.

\bibitem[{Nan et~al.(2020{\natexlab{a}})Nan, Guo, Sekulic, and
  Lu}]{nan-etal-2020-reasoning}
Guoshun Nan, Zhijiang Guo, Ivan Sekulic, and Wei Lu. 2020{\natexlab{a}}.
\newblock \href {https://doi.org/10.18653/v1/2020.acl-main.141} {Reasoning with
  latent structure refinement for document-level relation extraction}.
\newblock In \emph{Proceedings of the 58th Annual Meeting of the Association
  for Computational Linguistics}, pages 1546--1557, Online. Association for
  Computational Linguistics.

\bibitem[{Nan et~al.(2020{\natexlab{b}})Nan, Guo, Sekulic, and
  Lu}]{nan2020reasoning}
Guoshun Nan, Zhijiang Guo, Ivan Sekulic, and Wei Lu. 2020{\natexlab{b}}.
\newblock \href {https://doi.org/10.18653/v1/2020.acl-main.141} {Reasoning with
  latent structure refinement for document-level relation extraction}.
\newblock In \emph{Proceedings of the 58th Annual Meeting of the Association
  for Computational Linguistics}, pages 1546--1557, Online. Association for
  Computational Linguistics.

\bibitem[{Peng et~al.(2017)Peng, Poon, Quirk, Toutanova, and
  Yih}]{peng2017cross}
Nanyun Peng, Hoifung Poon, Chris Quirk, Kristina Toutanova, and Wen-tau Yih.
  2017.
\newblock \href {https://doi.org/10.1162/tacl_a_00049} {Cross-sentence n-ary
  relation extraction with graph {LSTM}s}.
\newblock \emph{Transactions of the Association for Computational Linguistics},
  5:101--115.

\bibitem[{Pennington et~al.(2014)Pennington, Socher, and Manning}]{glove}
Jeffrey Pennington, Richard Socher, and Christopher Manning. 2014.
\newblock \href {https://doi.org/10.3115/v1/D14-1162} {{G}lo{V}e: Global
  vectors for word representation}.
\newblock In \emph{Proceedings of the 2014 Conference on Empirical Methods in
  Natural Language Processing ({EMNLP})}, pages 1532--1543, Doha, Qatar.
  Association for Computational Linguistics.

\bibitem[{Quirk and Poon(2017)}]{quirk-poon-2017-distant}
Chris Quirk and Hoifung Poon. 2017.
\newblock \href {https://www.aclweb.org/anthology/E17-1110} {Distant
  supervision for relation extraction beyond the sentence boundary}.
\newblock In \emph{Proceedings of the 15th Conference of the {E}uropean Chapter
  of the Association for Computational Linguistics: Volume 1, Long Papers},
  pages 1171--1182, Valencia, Spain. Association for Computational Linguistics.

\bibitem[{Stevenson(2006)}]{stevenson2006fact}
Mark Stevenson. 2006.
\newblock Fact distribution in information extraction.
\newblock \emph{Language resources and evaluation}, 40(2):183--201.

\bibitem[{Swampillai and Stevenson(2010)}]{swampillai2010inter}
Kumutha Swampillai and Mark Stevenson. 2010.
\newblock \href
  {http://www.lrec-conf.org/proceedings/lrec2010/pdf/905_Paper.pdf}
  {Inter-sentential relations in information extraction corpora}.
\newblock In \emph{Proceedings of the Seventh International Conference on
  Language Resources and Evaluation ({LREC}'10)}, Valletta, Malta. European
  Language Resources Association (ELRA).

\bibitem[{Wu et~al.(2019)Wu, Luo, Leung, Ting, and Lam}]{wu2019renet}
Ye~Wu, Ruibang Luo, Henry~CM Leung, Hing-Fung Ting, and Tak-Wah Lam. 2019.
\newblock Renet: A deep learning approach for extracting gene-disease
  associations from literature.
\newblock In \emph{International Conference on Research in Computational
  Molecular Biology}, pages 272--284. Springer.

\bibitem[{Yao et~al.(2019)Yao, Ye, Li, Han, Lin, Liu, Liu, Huang, Zhou, and
  Sun}]{yao_docred_2019}
Yuan Yao, Deming Ye, Peng Li, Xu~Han, Yankai Lin, Zhenghao Liu, Zhiyuan Liu,
  Lixin Huang, Jie Zhou, and Maosong Sun. 2019.
\newblock \href {https://doi.org/10.18653/v1/P19-1074} {{D}oc{RED}: A
  large-scale document-level relation extraction dataset}.
\newblock In \emph{Proceedings of the 57th Annual Meeting of the Association
  for Computational Linguistics}, pages 764--777, Florence, Italy. Association
  for Computational Linguistics.

\bibitem[{Zeng et~al.(2020{\natexlab{a}})Zeng, Xu, Chang, and
  Li}]{zeng-etal-2020-double}
Shuang Zeng, Runxin Xu, Baobao Chang, and Lei Li. 2020{\natexlab{a}}.
\newblock \href {https://doi.org/10.18653/v1/2020.emnlp-main.127} {Double graph
  based reasoning for document-level relation extraction}.
\newblock In \emph{Proceedings of the 2020 Conference on Empirical Methods in
  Natural Language Processing (EMNLP)}, pages 1630--1640, Online. Association
  for Computational Linguistics.

\bibitem[{Zeng et~al.(2020{\natexlab{b}})Zeng, Xu, Chang, and
  Li}]{zeng2020double}
Shuang Zeng, Runxin Xu, Baobao Chang, and Lei Li. 2020{\natexlab{b}}.
\newblock \href {https://doi.org/10.18653/v1/2020.emnlp-main.127} {Double graph
  based reasoning for document-level relation extraction}.
\newblock In \emph{Proceedings of the 2020 Conference on Empirical Methods in
  Natural Language Processing (EMNLP)}, pages 1630--1640, Online. Association
  for Computational Linguistics.

\bibitem[{Zhang et~al.(2018)Zhang, Qi, and Manning}]{zhang-etal-2018-graph}
Yuhao Zhang, Peng Qi, and Christopher~D. Manning. 2018.
\newblock \href {https://doi.org/10.18653/v1/D18-1244} {Graph convolution over
  pruned dependency trees improves relation extraction}.
\newblock In \emph{Proceedings of the 2018 Conference on Empirical Methods in
  Natural Language Processing}, pages 2205--2215, Brussels, Belgium.
  Association for Computational Linguistics.

\end{thebibliography}

\end{document}